\documentclass{article}

    \usepackage[final, nonatbib]{neurips_2020}

\usepackage[utf8]{inputenc} % allow utf-8 input
\usepackage[T1]{fontenc}    % use 8-bit T1 fonts
\usepackage{hyperref}       % hyperlinks
\usepackage{url}            % simple URL typesetting
\usepackage{booktabs}       % professional-quality tables
\usepackage{amsfonts}       % blackboard math symbols
\usepackage{nicefrac}       % compact symbols for 1/2, etc.
\usepackage{microtype}      % microtypography
\usepackage{todonotes}
\usepackage{comment}
\usepackage{subcaption}
\usepackage{amsmath}
\usepackage[
backend=biber,
style=numeric,
citestyle=numeric
]{biblatex}
\DeclareMathOperator*{\argmin}{arg\,min}

\title{Mask-Guided Discovery of Semantic Manifolds in Generative Models}
\author{%
  Mengyu Yang \\
   \footnotesize\texttt{my.yang@mail.utoronto.ca} \\
  \footnotesize{BMO Lab for Creative Research} \\
   University of Toronto \\
  \And
   David Rokeby \\
   \footnotesize\texttt{david.rokeby@utoronto.ca} \\
  \footnotesize{BMO Lab for Creative Research}\\
   University of Toronto \\
    \And
   Xavier Snelgrove \\
    \footnotesize\texttt{xavier@cs.toronto.edu} \\
  \footnotesize{BMO Lab for Creative Research}\\
   University of Toronto \\
 }
 
\begin{document}

\maketitle

\section{Introduction}
Advances in the realm of Generative Adversarial Networks (GANs) \cite{goodfellow_generative_2014} have led to architectures capable of producing amazingly realistic images such as StyleGAN2  \cite{karras_analyzing_2020}, which, when trained on the FFHQ dataset \cite{karras_style-based_2019}, generates images of human faces from random vectors in a lower-dimensional latent space. Unfortunately, this space is entangled -- translating a latent vector along its axes does not correspond to a meaningful transformation in the output space  (e.g., smiling mouth, squinting eyes). The model behaves as a black box, providing neither control over its output nor insight into the structures it has learned from the data. 

%\footnote{For concreteness we will use the example of a face-generating network throughout, but the methods generalize to other classes of image as well. Refer to supplementary material for more}
However, the smoothness of the mappings from latent vectors to faces plus empirical evidence \cite{shen_interpreting_2020,creswell_inverting_2018} suggest that manifolds of meaningful transformations are in fact hidden inside the latent space but obscured by not being axis-aligned or even linear. Travelling along these manifolds would provide puppetry-like abilities to manipulate faces while studying their geometry would provide insight into the nature of the face variations present in the dataset -- revealing and quantifying the degrees-of-freedom of eyes, mouths, \emph{etc}.

We present a method to explore the manifolds of changes of spatially localized regions of the face. Our method discovers smoothly varying sequences of latent vectors along these manifolds suitable for creating animations. Unlike existing disentanglement methods that either require labelled data \cite{shen_interpreting_2020,wei_maggan_2020} or explicitly alter internal model parameters \cite{alharbi_disentangled_2020,broad_network_2020}, our method is an optimization-based approach guided by a custom loss function and manually defined region of change. Our code is open-sourced, which can be found, along with supplementary results, on our project page\footnote{\url{https://github.com/bmolab/masked-gan-manifold}}. 

\section{Method}

We design functions defined on the images generated by our pre-trained model (we will continue to work with the example of StyleGAN2 trained on FFHQ). The desired property of these functions is that they are at their minimum when only the target region of the face (for instance, the mouth) has changed. We then use standard optimization techniques to discover smoothly varying paths through the latent space that lie on the manifold.

We start with a user-provided initial generated image $\vec{x}^* = G(\vec{z}^*)$, where $G$ is the generator network and $\vec{z}^*$ some latent vector (note that in this work we use StyleGAN2's higher-dimensional intermediate latent space $\vec{z}^* \in \mathcal{W+}$, refer to \cite{karras_style-based_2019} for details). We then define a rectangular mask region $M$ over the image, for instance around the mouth, and define $\vec{x}^*_{M}$ as the image formed by cropping $\vec{x}^*$ to $M$, and $\vec{x}^*_{\overline{M}}$ as its complement (i.e. the rest of the image). We seek a manifold containing 
images which have primarily changed in the mouth region $M$ but not in the rest of the image $\overline{M}$. We can define this manifold as minima of the function
\begin{equation}
    \mathcal{L}_X(\vec{x}^*,\vec{x}_i; M) = |D(\vec{x}^*_M, \vec{x}_{i,M})-c| + D(\vec{x}^*_{\overline{M}},\vec{x}_{i,\overline{M}})
\end{equation}
where $D(\cdot, \cdot)$ is a distance function between images. We have experimented with both $L^2$ pixel-wise distance and the LPIPS perceptual loss \cite{zhang_unreasonable_2018}. $\mathcal{L}_X$ satisfies our requirement as it is minimal when the target region has changed by a factor of $c$ while the rest of the image remains unchanged.

In order to create smoothly varying animations that explore this manifold, we use a physically-inspired model of masses connected by springs. Take a matrix of $n$ latent vectors $Z = [\vec{z}_1, \ldots, \vec{z}_n]^T \in \mathbb{R}^{n \times w}$ where $w$ is the dimension of the latent space. The vectors are connected by springs of rest-length $\sigma$ (an adjustable parameter) in series, encouraging each to be similar, but not too similar, to its neighbours. We further encourage the path to have minimal curvature by also adding higher-order ``stiffener'' springs connecting to vectors that are further apart. This system can be formalized as follows,
\begin{equation}
    \mathcal{L}_{spring}(Z;k) = \sum_{i=1}^{n-k} \left( || \vec{z}_i - \vec{z}_{i+k} ||_2 - k\,\sigma \right)^2
\end{equation}

For our experiments, we include $k=1,2$. Putting everything together, given a reference latent vector $\vec{z}^*$ and mask region $M$, we use the L-BFGS \cite{liu_limited_1989} algorithm to optimize and find $\tilde{Z}$, as seen in Equation \ref{final_eqn}, where $\alpha, \beta, \gamma$ are tuneable parameters controlling the importance of each term. The result of this optimization is visually represented in Figure \ref{fig:image_grid}, left. 

\begin{equation}\label{final_eqn}
    \tilde{Z} = \argmin_{Z}\; \alpha\sum_{i=1}^n\mathcal{L}_X(G(\vec{z}^*), G(\vec{z}_i); M) + \beta \mathcal{L}_{spring}(Z; 1) + \gamma \mathcal{L}_{spring}(Z; 2)
\end{equation}

\section{Results and discussion}

% \begin{figure}[t]
%   \centering
%  \begin{subfigure}{.49\textwidth}
%     \centering
%     \def\svgwidth{\linewidth}
% {\scriptsize\input{Landscape.pdf_tex}}
% %\caption{}
% \end{subfigure}\hfill
%  \begin{subfigure}{.49\textwidth}
%   \centering
%   \includegraphics[width=\linewidth]{Figures/three_faces.jpg}
%   %\caption{}
% \end{subfigure}
%   \caption{\textbf{Left}: A choice of seed vector $\vec{z}^*$, shown as an orange dot, and mask region $M$ creates a function $\mathcal{L}_X$ illustrated here on the latent space of the generative model. The optimization seeks a set of vectors $\{\vec{z}_i\}$, shown as red dots, that lie along the minima of this landscape. They are encouraged to be evenly spaced along a path with minimal curvature by the spring loss $\mathcal{L}_{spring}$, where springs of order $k=1, 2$ are shown as green and blue connectors respectively. \textbf{Right}: Various results of our algorithm. Each row consists of a different seed vector and mask region as shown in the first column. The other columns are selected images from the generated sequences $\{\vec{z}_i\}$. Note that the change in the images is well localized to the masked region. Also note that we use a large value of $c$ to exaggerate the changes for clarity. Refer to the Appendix for many more results.} 
% \label{fig:image_grid}
% \end{figure}

\begin{figure}[t]
  \centering
 \begin{subfigure}{.49\textwidth}
    \centering
    \includegraphics[width=\linewidth]{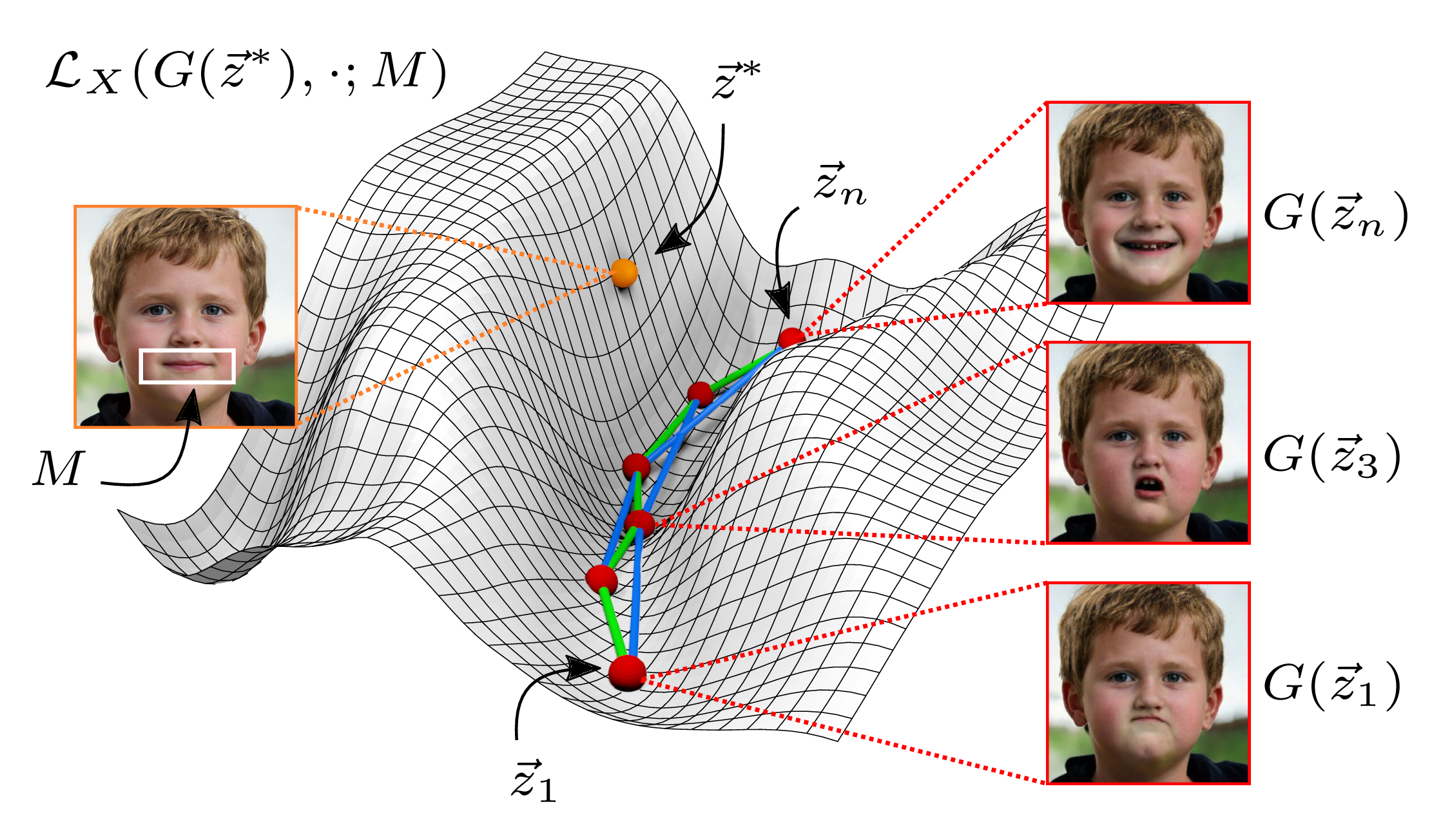}
    %\caption{}
\end{subfigure}\hfill
 \begin{subfigure}{.49\textwidth}
  \centering
  \includegraphics[width=\linewidth]{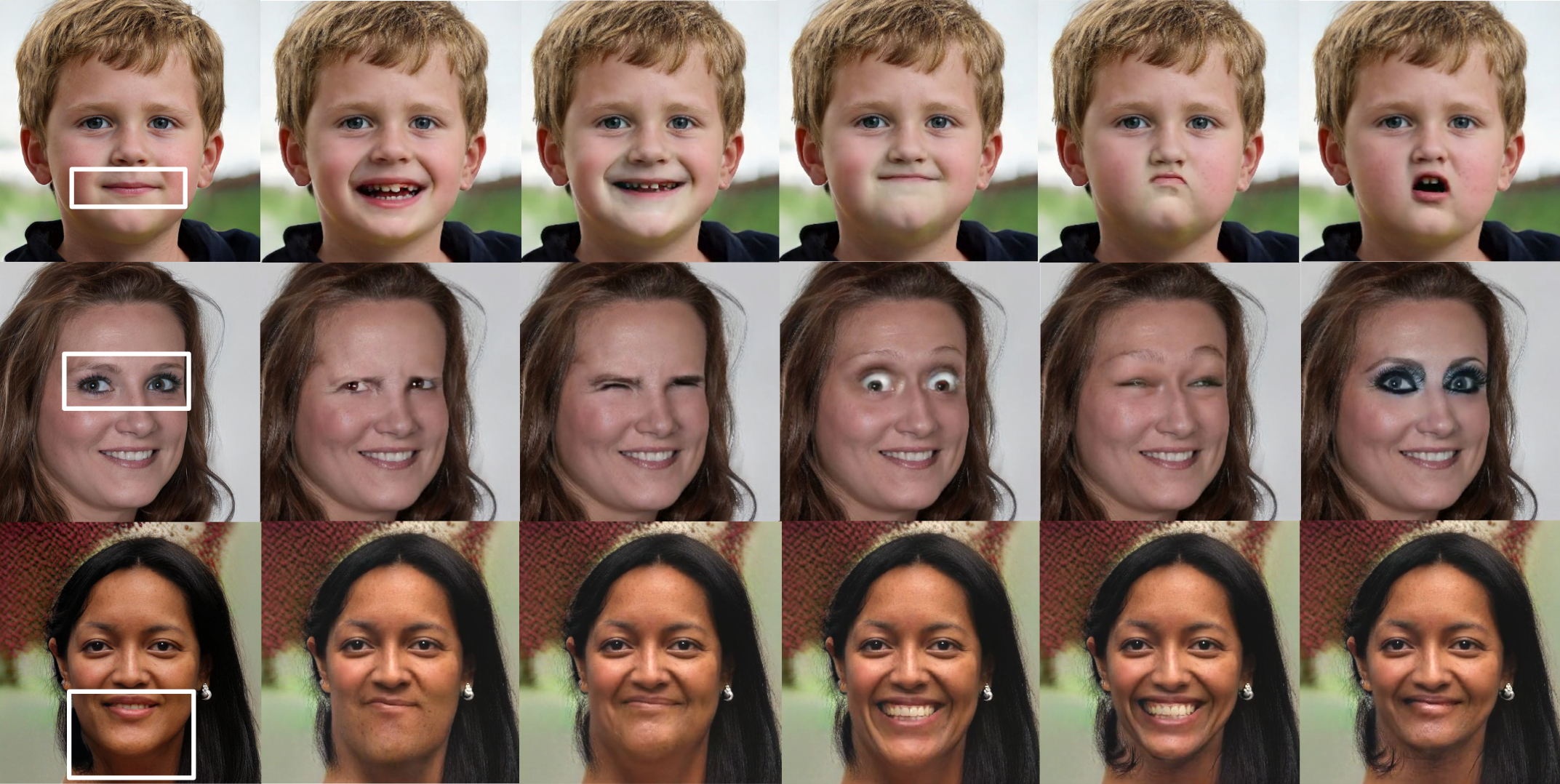}
  %\caption{}
\end{subfigure}
  \caption{\textbf{Left}: A choice of seed vector $\vec{z}^*$, shown as an orange dot, and mask region $M$ creates a function $\mathcal{L}_X$ illustrated here on the latent space of the generative model. The optimization seeks a set of vectors $\{\vec{z}_i\}$, shown as red dots, that lie along the minima of this landscape. They are encouraged to be evenly spaced along a path with minimal curvature by the spring loss $\mathcal{L}_{spring}$, where springs of order $k=1, 2$ are shown as green and blue connectors respectively. \textbf{Right}: Various results of our algorithm. Each row consists of a different seed vector and mask region as shown in the first column. The other columns are selected images from the generated sequences $\{\vec{z}_i\}$. Note that the change in the images is well localized to the masked region. Also note that we use a large value of $c$ to exaggerate the changes for clarity. Refer to the Appendix for many more results.} 
\label{fig:image_grid}
\end{figure}

Figure \ref{fig:image_grid}, right, shows some of our experimental results. It can be seen that changes to the face are all localized within the mask region while minimal change occurs outside. More importantly, we demonstrate that our method is generalizable to any mask region of choice as well as initial face (see Figures \ref{fig:boy_mouth} to \ref{fig:woman_mouth} in the Appendix for additional experiments). The spring constraints of our method are designed to generate smooth videos, please refer to our supplementary material to experience this qualitatively.

This work is a small contribution towards the larger vision of exploring and characterizing the semantic manifolds lurking in the latent spaces of generative models. Generalizing to different models, different dimensionalities of manifolds, and more controls than just rectangular masks are a small sampling of the natural extensions of this line of inquiry.

\section{Ethical implications}
The StyleGAN2 model we use is capable of generating realistic faces while also demonstrating a proficient understanding of how faces tend to vary in the dataset. Given these qualities, GANs can be used as a popular tool for modelling and promulgating what is considered to be ``normal'', which if used uncritically, could marginalize people labelled ``abnormal'' by these systems \cite{crawford_trouble_2017}.

Furthermore, there has been much popular discussion about whether we are entering a post-truth contemporary era, where generative tools such as the one we present here have raised fears of hyper-realistic ``deepfake'' videos impersonating real people, poisoning the information ecology and further eroding trust in any consensus reality \cite{vincent_watch_2018}.

Perhaps more subtly, our method and others like it can create very physically plausible videos of faces changing in ``unnatural'' ways, such as shifting bone structure, smoothly varying a face from one identity to another. If such videos become commonplace in our culture, might this contribute to a reconfiguring of our traditional conception of separate, fixed, and individual identities towards fluid, overlapping, and changeable ones? The consequences of such a fundamental shift, be they negative, positive, or neutral are difficult to anticipate but worthy of consideration.

\printbibliography

\newpage
\section{Appendix}

Below are some additional figures of experiments with different faces and masks. For all figures below, the first column is the reference image $\vec{x}^*$ with mask region shown. Unless its value is explicitly stated, we use a large value of $c$ to exaggerate the changes for clarity. 

We encourage readers to view our video animations, from which stills were taken to create these figures below. An important aspect of our method is that it creates smooth animations while exploring the manifolds. As a result, the video animations convey much more visual information, whereas some of that is lost with still figures. Refer to supplementary materials for the animations.

\begin{figure}[ht]
     \centering
     \begin{subfigure}{\textwidth}
         \centering
         \includegraphics[width=\textwidth]{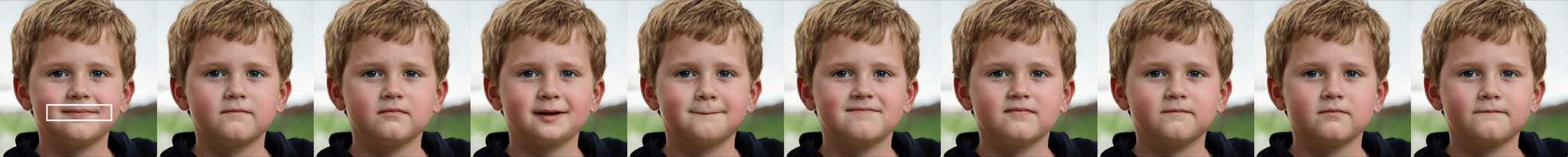}
         \caption{$c=0.15$}
         \label{fig:boy_mouth_0.15}
     \end{subfigure}
     \hfill
     \begin{subfigure}{\textwidth}
         \centering
         \includegraphics[width=\textwidth]{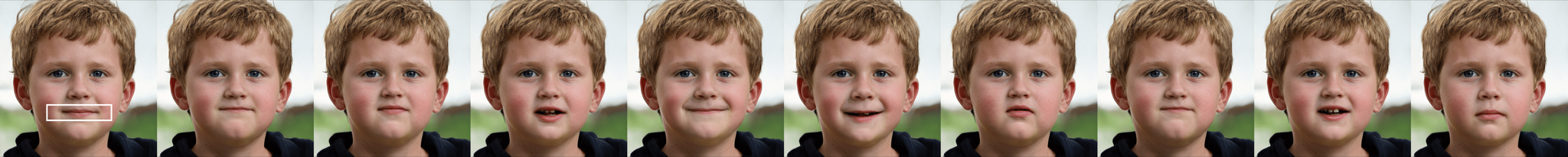}
         \caption{$c=0.25$}
         \label{fig:boy_mouth_0.25}
     \end{subfigure}
     \hfill
     \begin{subfigure}{\textwidth}
         \centering
         \includegraphics[width=\textwidth]{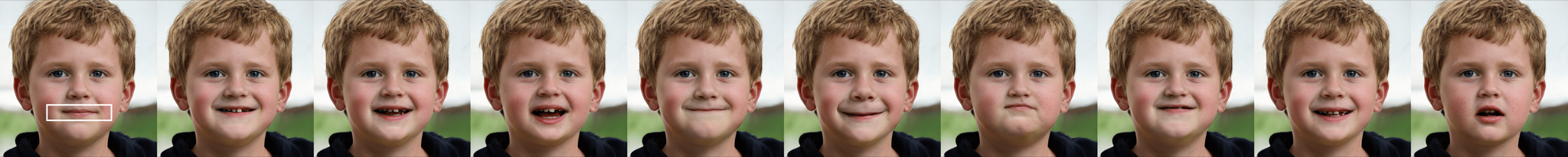}
         \caption{$c=0.35$}
         \label{fig:boy_mouth_0.35}
     \end{subfigure}
     \hfill
     \begin{subfigure}{\textwidth}
         \centering
         \includegraphics[width=\textwidth]{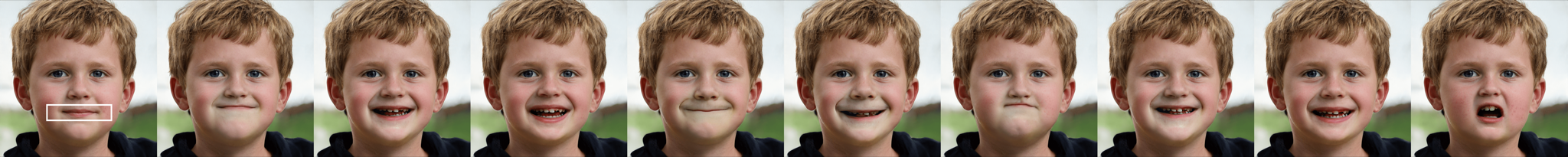}
         \caption{$c=0.45$}
         \label{fig:boy_mouth_0.45}
     \end{subfigure}
        \caption{Each row represents an experiment with a different offset $c$ value on the same face and mask region.}
        \label{fig:boy_mouth}
\end{figure}

\begin{figure}[ht]
    \centering
    \includegraphics[width=\textwidth]{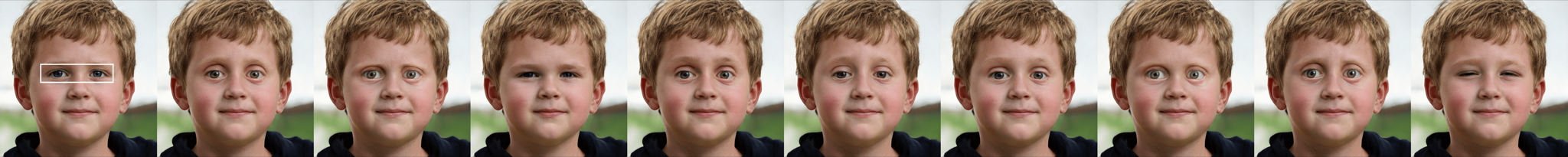}
    \caption{Experiment with a single offset $c$ value with mask region around the eyes.}
    \label{fig:boy_eyes}
\end{figure}

\begin{figure}[ht]
    \centering
    \includegraphics[width=\textwidth]{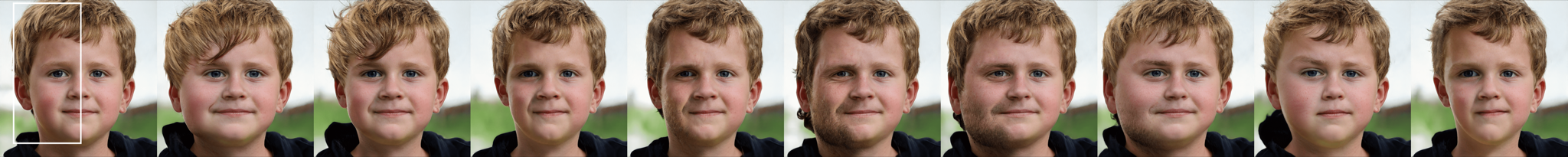}
    \caption{Experiment with a single offset $c$ value with mask region around the right half of the face.}
    \label{fig:boy_half_face}
\end{figure}

\begin{figure}[ht]
    \centering
    \includegraphics[width=\textwidth]{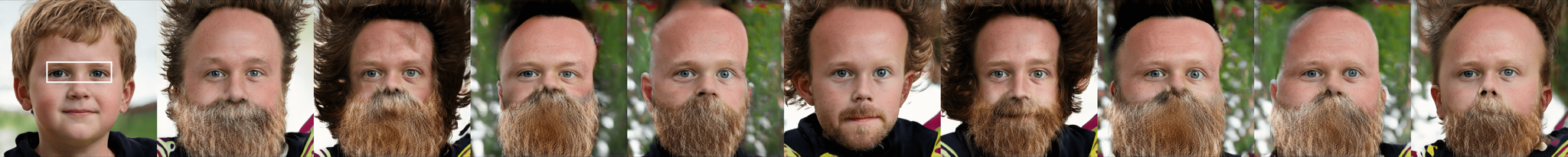}
    \caption{Experiment with a single offset $c$ value with mask region around everything except the eye region (i.e., eye region remains unchanged).}
    \label{fig:boy_non_eye}
\end{figure}

\begin{figure}[ht]
    \centering
    \includegraphics[width=\textwidth]{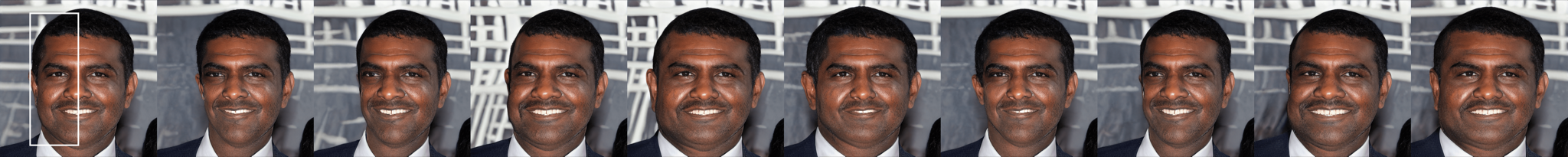}
    \caption{Experiment with a single offset $c$ value with mask region around the right half of the face.}
    \label{fig:man_half_face}
\end{figure}

\begin{figure}[ht]
    \centering
    \includegraphics[width=\textwidth]{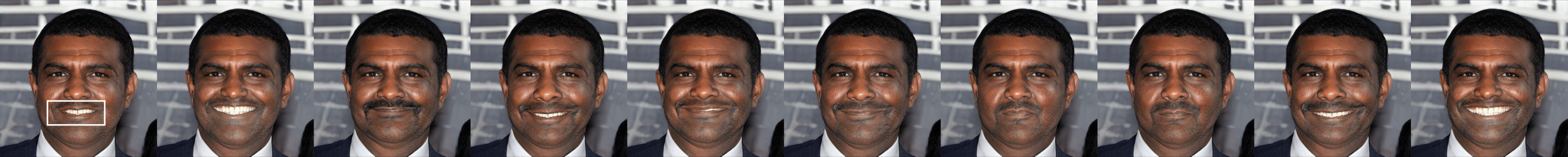}
    \caption{Experiment with a single offset $c$ value with mask region around the mouth.}
    \label{fig:man_mouth}
\end{figure}

\begin{figure}[ht]
    \centering
    \includegraphics[width=\textwidth]{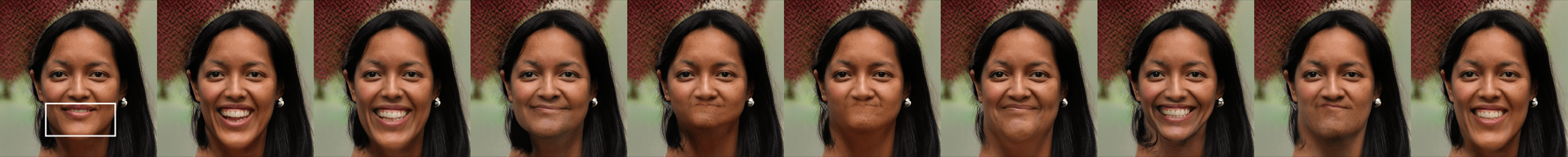}
    \caption{Experiment with a single offset $c$ value with mask region around the mouth and chin.}
    \label{fig:woman2_mouth_chin}
\end{figure}

\begin{figure}[ht]
    \centering
    \includegraphics[width=\textwidth]{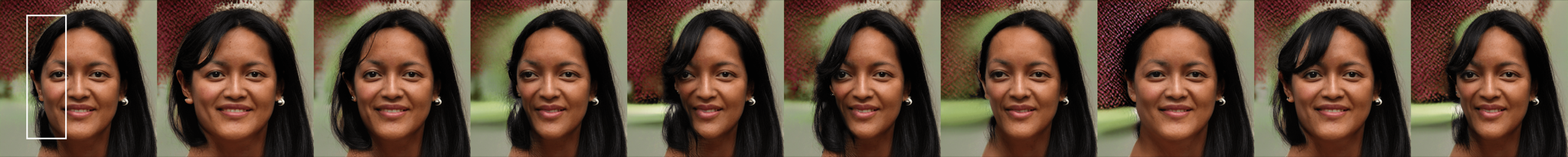}
    \caption{Experiment with a single offset $c$ value with mask region around the right half of the face.}
    \label{fig:woman2_half_face}
\end{figure}

\begin{figure}[ht]
    \centering
    \includegraphics[width=\textwidth]{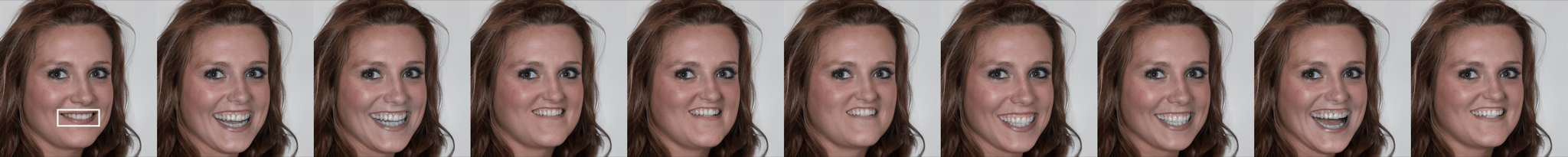}
    \caption{Experiment with a single offset $c$ value with mask region around the mouth.}
    \label{fig:woman_mouth}
\end{figure}

\end{document}